\documentclass[runningheads]{llncs}
\usepackage{colortbl}
\usepackage{xcolor}
\usepackage{graphicx} 
\usepackage{adjustbox}
\usepackage{amsmath}
\usepackage{float}
\usepackage{makecell}
\usepackage{amssymb}
\usepackage{textcomp}
\usepackage{float}
\usepackage[colorlinks=true, linkcolor=blue, citecolor=blue, urlcolor=blue]{hyperref}  
\usepackage[T1]{fontenc}
\usepackage{orcidlink}
\usepackage{marvosym}
\usepackage{graphicx}
\begin{document}
\title{GK-SMOTE: A Hyperparameter-free Noise-Resilient Gaussian KDE-Based Oversampling Approach}
%
%
\author{Mahabubur Rahman Miraj\inst{1}\orcidlink{0000-0001-5591-3032} \and
Hongyu Huang\inst{1}\orcidlink{0000-0001-9052-4868} \and
Ting Yang\inst{1}\orcidlink{0009-0006-0410-7376}\and
Jinxue Zhao\inst{1}\orcidlink{0009-0007-3449-8563} \and
Nankun Mu\inst{1}\textsuperscript{(\Letter)}\orcidlink{0009-0005-1560-8267}\and
Xinyu Lei\inst{2}\orcidlink{0000-0001-8799-7875}}
\authorrunning{M.R. Miraj, Hongyu Huang, et al.}
%
\institute{College of Computer Science, Chongqing University, Chongqing 400044, China\\ \and
Department of Computer Science, Michigan Technological University, Houghton, MI 49931, Michigan, United States\\
\email{nankun.mu@cqu.edu.cn; miraj@stu.cqu.edu.cn}}
\maketitle              
\vspace{-6mm}

\begin{abstract}
Imbalanced classification is a significant challenge in machine learning, especially in critical applications like medical diagnosis, fraud detection, and cybersecurity. Traditional oversampling techniques, such as SMOTE, often fail to handle label noise and complex data distributions, leading to reduced classification accuracy. In this paper, we propose GK-SMOTE, a hyperparameter-free, noise-resilient extension of SMOTE, built on Gaussian Kernel Density Estimation (KDE). GK-SMOTE enhances class separability by generating synthetic samples in high-density minority regions, while effectively avoiding noisy or ambiguous areas. This self-adaptive approach uses Gaussian KDE to differentiate between safe and noisy regions, ensuring more accurate sample generation without requiring extensive parameter tuning. Our extensive experiments on diverse binary classification datasets demonstrate that GK-SMOTE outperforms existing state-of-the-art oversampling techniques across key evaluation metrics, including MCC, Balanced Accuracy, and AUPRC. The proposed method offers a robust, efficient solution for imbalanced classification tasks, especially in noisy data environments, making it an attractive choice for real-world applications. 
\end{abstract}
\section{Introduction}
Imbalanced classification has emerged as a prominent challenge in data mining and machine learning due to the uneven distribution of instances across classes in real-world datasets \cite{liu2025graphsurvey}. In such cases, the majority class dominates, while the minority class, which often contains more significant instances, is underrepresented. This phenomenon is particularly prevalent in critical applications such as medical diagnostics, network intrusion detection, and bankruptcy prediction \cite{chen2024survey}. Traditional classification methods tend to optimize for overall accuracy by minimizing loss functions, which often results in a bias towards the majority class. In datasets with a large majority-to-minority class ratio (e.g., 99:1), even a simple classifier that labels all instances as belonging to the majority class can achieve an artificially high accuracy of 99\%. However, this leads to poor performance on the minority class, which is crucial in many practical contexts, such as medical diagnoses, where the identification of patients with a disease is far more important than identifying non-diseased individuals \cite{liu2025graphsurvey}.

To address these challenges, various strategies have been developed, falling into three primary categories: data-level methods, algorithm-level methods, and cost-sensitive methods. Among these, data-level methods, particularly oversampling techniques, have gained significant traction due to their simplicity and adaptability \cite{zhang2023density}. Oversampling increases the representation of the minority class without discarding valuable data from the majority class, unlike under-sampling techniques. Additionally, oversampling methods, such as SMOTE (Synthetic Minority Over-sampling Technique), have been found to perform better than under-sampling when evaluated using metrics like the area under the ROC curve (AUC). SMOTE creates synthetic samples by interpolating between adjacent instances in the feature space, enhancing classifier performance by balancing the class distribution. However, SMOTE has notable limitations, especially when dealing with noisy data \cite{chawla2002smote}.

Recent research indicates that the imbalance ratio alone is insufficient to explain the deterioration in classification performance; factors such as data complexity and label noise play more significant roles \cite{carvalho2025resampling}. Label noise, which refers to inaccurately labeled instances, is a common issue in real-world datasets \cite{cinar2025synergistic}. SMOTE’s interpolation approach can exacerbate the negative effects of label noise, especially in the regions near decision boundaries, by generating noisy and overlapping samples. This reduces the separability between classes and harms classifier performance \cite{chawla2002smote}. Therefore, there is an urgent need for more robust oversampling techniques that address both label noise and complex data distributions. In response, this work introduces GK-SMOTE, a self-adaptive, Gaussian Kernel Density Estimation (KDE)-based robust version of SMOTE. GK-SMOTE aims to improve classification performance in imbalanced datasets with label noise. Key innovations include:
\vspace{-6.5mm}

\begin{itemize}
    \item Using Gaussian KDE to assess the local density of minority samples and categorize them into safe and borderline areas. This helps in generating synthetic samples that improve class separability and mitigate label noise.
    \item Adapting the density-based approach to focus on high-density regions of the minority class, ensuring that synthetic samples are created in areas where they enhance classification performance, while avoiding noisy regions.
    \item Maintaining simplicity and efficiency by leveraging the same k-nearest neighbors used in KDE for sample generation, facilitating integration into existing workflows without the need for additional parameters.
    \item Extensive validation on various binary classification benchmark datasets with varying degrees of label noise, demonstrating the superior performance of GK-SMOTE over other advanced oversampling techniques based on evaluation metrics such as MCC (Matthews Correlation Coefficient), Balanced Accuracy (BAc), and AUPRC (Area Under Precision-Recall Curve).
\end{itemize}
By targeting label noise and improving class separability, GK-SMOTE offers a robust solution for handling imbalanced datasets, yielding more reliable classification of minority class instances.

\section{Related Work}
\subsection{State of the Art Oversampling Methods}
Over-sampling techniques such as SMOTE and its variants have become fundamental for addressing imbalanced datasets in machine learning. SMOTE generates synthetic minority class samples through interpolation between neighboring instances, improving class balance, but struggles with noisy data and label inaccuracies, often resulting in reduced performance when label noise is present \cite{chawla2002smote}. Borderline-SMOTE extends SMOTE by focusing on the decision boundary and generating synthetic samples in regions near the class boundary, which is particularly effective for classes that are not easily separable but still falls short when label noise is present, as it does not explicitly address noise robustness \cite{han2005borderline}. ADASYN (Adaptive Synthetic Sampling) further extends SMOTE by adaptively generating samples in low-density regions to improve classifier performance by focusing on difficult-to-classify instances, but similarly suffers from issues when noisy labels are present, as it does not fully address label noise \cite{he2008adasyn}. AB-SMOTE (Adaptive Borderline-SMOTE) enhances SMOTE by introducing a weighting mechanism that focuses on minority class instances near the decision boundary, improving data complexity handling and class separability but not fully addressing label noise \cite{majzoub2020absmote}. Safe-Level-SMOTE filters safe regions for synthetic sample generation to mitigate the adverse effects of noise but still lacks noise robustness and struggles in environments with high label noise, limiting its applicability in real-world scenarios where label noise is common \cite{bunkhumpornpat2009safelevel}. Km-SMOTE improves SMOTE by using clustering methods such as K-means to generate synthetic samples based on centroids, aiding class separability and data complexity, but it does not directly address the problem of label noise, which may result in generating noisy samples in noisy environments \cite{guo2019improved}. Despite these advancements, these methods still fail to fully address label noise, limiting their robustness in noisy environments.

\subsection{Gaussian-Based SMOTE Methods}
Gaussian-based oversampling methods, such as GMF-SMOTE \cite{xu2022gmm}, NGOS \cite{shao2023noise}, MGD \cite{xie2023instance}, HGDO \cite{jia2024hgdo}, and GDO \cite{xie2020gaussian}, have been developed to improve noise robustness and sampling efficiency. GMF-SMOTE utilizes Gaussian mixture models to filter out noisy data before generating synthetic samples, improving performance in noisy environments by focusing on areas with better class separability and reducing the risk of generating noisy samples \cite{xu2022gmm}. NGOS (noise gaussian oversampling) \cite{shao2023noise} and MGD (multivariate gaussian distribution)  also use Gaussian distributions to avoid generating samples in noisy regions, effectively handling noisy environments to some extent \cite{xie2023instance}. However, these methods still require hyperparameter tuning, which can be computationally expensive and inefficient, particularly in high-dimensional settings. HGDO (hypergraph Gaussian distribution oversampling) \cite{jia2024hgdo} and GDO aim to leverage Gaussian distributions for more robust noise handling, showing better performance in complex data environments compared to traditional methods \cite{xie2020gaussian}. However, they still suffer from the need for hyperparameter tuning, making them less user-friendly and harder to integrate into practical applications, with their computational inefficiency in high-dimensional spaces being a limiting factor in real-world use cases. Among these methods, GK-SMOTE stands out as the most robust solution to label noise, as it leverages Gaussian Kernel Density Estimation (KDE) to focus on high-density regions of the minority class while avoiding noisy areas. Unlike GMF-SMOTE \cite{xu2022gmm} and NGOS \cite{shao2023noise}, GK-SMOTE is fully hyperparameter-free, making it more user-friendly and adaptable to practical applications. This unique feature ensures that GK-SMOTE strikes a balance between robustness, efficiency, and simplicity, enabling it to handle imbalanced datasets with label noise effectively, and providing a more reliable and efficient solution for real-world imbalanced classification problems, particularly in noisy environments where other methods may struggle as shown in the Table: \ref{table: 1}.

\begin{table}[h!]
\centering
\caption{Comparison of Oversampling Methods with GK-SMOTE}
\scriptsize  
\renewcommand{\arraystretch}{0.9}  
\setlength{\tabcolsep}{0.9pt}  
\begin{tabular}{l c c c c c c}
\hline
\makecell{Oversampling\\Methods} & \makecell{Noise\\Robustness} & \makecell{Handle \\ Label Noise} & \makecell{Data\\Complexity} & \makecell{Compute \\ Efficiency}  & \makecell{Hyperparameter \\ Methods} & \makecell{Gaussian KDE \\Based Methods} \\
\hline
SMOTE \cite{chawla2002smote} & \texttimes & \texttimes & \texttimes & \texttimes & \texttimes & \texttimes \\
Borderline-SMOTE \cite{han2005borderline} & \texttimes & \texttimes & \texttimes & \texttimes & \texttimes & \texttimes \\
ADASYN \cite{he2008adasyn} & \texttimes & \texttimes & \checkmark & \texttimes & \texttimes & \texttimes \\
AB-SMOTE \cite{majzoub2020absmote} & \checkmark & \texttimes & \checkmark & \checkmark & \texttimes & \texttimes \\
Km-SMOTE \cite{guo2019improved} & \checkmark & \texttimes & \checkmark & \texttimes & \texttimes & \texttimes \\
Safe-Level-SMOTE \cite{bunkhumpornpat2009safelevel} & \checkmark & \texttimes & \checkmark & \texttimes & \texttimes & \texttimes \\
GMF-SMOTE \cite{xu2022gmm} & \checkmark & \texttimes & \checkmark & \checkmark & \texttimes & \texttimes \\
NGOS \cite{shao2023noise} & \checkmark & \texttimes & \checkmark & \checkmark & \checkmark & \texttimes \\
MGD \cite{xie2023instance} & \checkmark & \checkmark & \checkmark & \checkmark & \checkmark & \texttimes \\
HGDO \cite{jia2024hgdo} & \checkmark & \checkmark & \checkmark & \checkmark & \checkmark & \texttimes \\
GDO \cite{xie2020gaussian} & \checkmark & \checkmark & \checkmark & \checkmark & \checkmark & \texttimes \\
GK-SMOTE & \checkmark & \checkmark & \checkmark & \checkmark & \checkmark & \checkmark \\
\hline
\label{table: 1}
\end{tabular}
\end{table}
\vspace{-10mm}

\section{Proposed Framework of GK-SMOTE}
\begin{figure}[h!]
    \centering
    \includegraphics[width=1.1\textwidth]{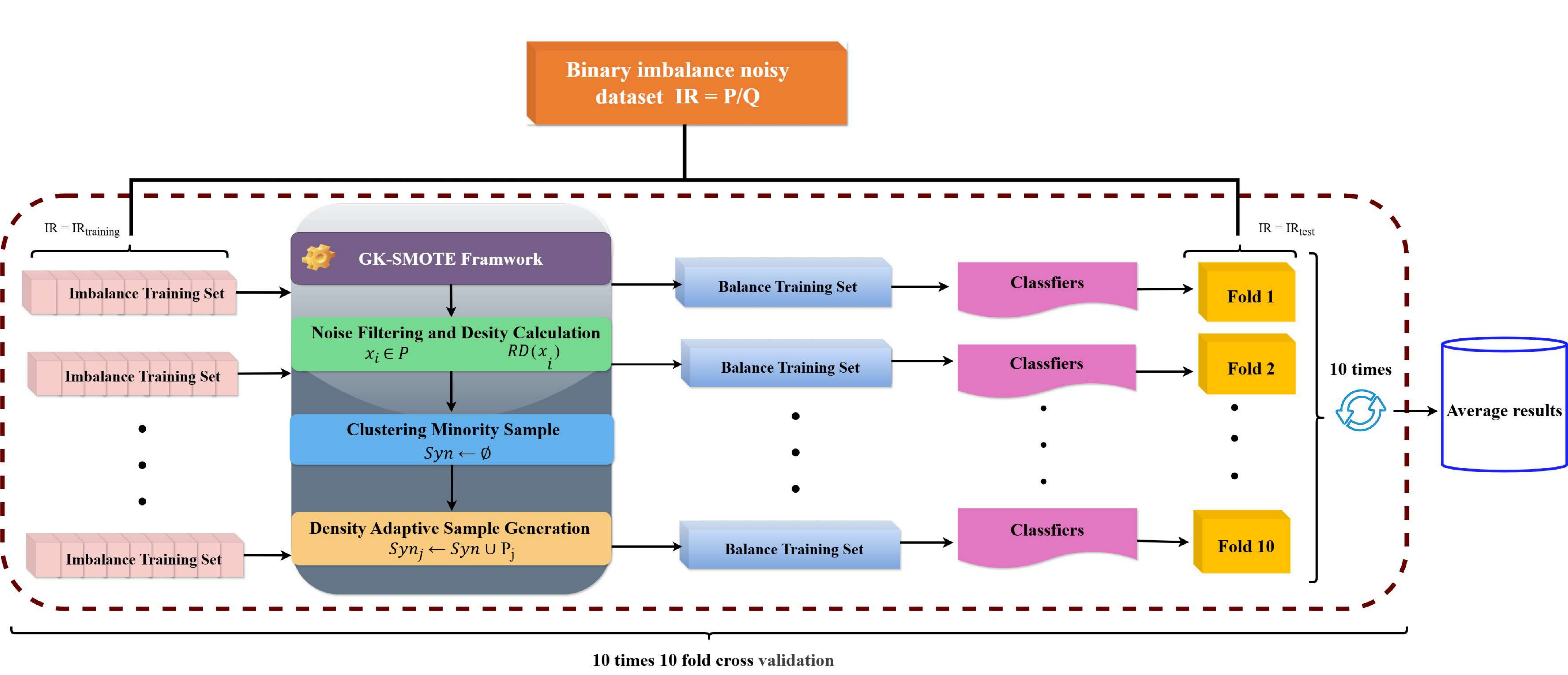} 
    \caption{GK-SMOTE Framework}
    \label{figure 1}
\end{figure}

We propose GK-SMOTE, a robust and adaptive extension of the traditional SMOTE (Synthetic Minority Over-sampling Technique), designed to address the challenges of class imbalance and label noise in machine learning. By leveraging Gaussian Kernel Density Estimation (KDE), GK-SMOTE creates synthetic data that avoids the common pitfalls of standard SMOTE, such as introducing noise or class overlap. The GK-SMOTE framework, outlined in Figure \ref{figure 1}, follows a structured process to generate synthetic samples while enhancing class separability. It starts with processing the imbalanced training set and calculating the class imbalance ratio (IR) between the majority class (Q) and the minority class (P). A noise filtering step is applied, followed by density calculation to differentiate minority samples based on their location within the dataset, helping to distinguish those near decision boundaries from those in safe regions. The framework then clusters the minority samples into borderline and safe categories using 2-means clustering, ensuring a targeted oversampling process. Synthetic samples are generated through interpolation using the k-nearest neighbors (k-NN) algorithm, and further refined with Gaussian KDE, focusing on high-density regions of the minority class to improve class separability. To evaluate the generated synthetic data, GK-SMOTE uses 10-fold cross-validation, ensuring robust performance across multiple folds, with results averaged for final performance metrics. This method operates hyperparameter-free, eliminating the need for manual tuning of parameters and ensuring computational efficiency by using the same k-NN algorithm for both clustering and oversampling. 
\vspace{-8mm}
\vspace{-10mm}

\subsection{GK-SMOTE Algorithm}
\begin{figure}[h!]
    \centering
    \includegraphics[width=1.0\textwidth]{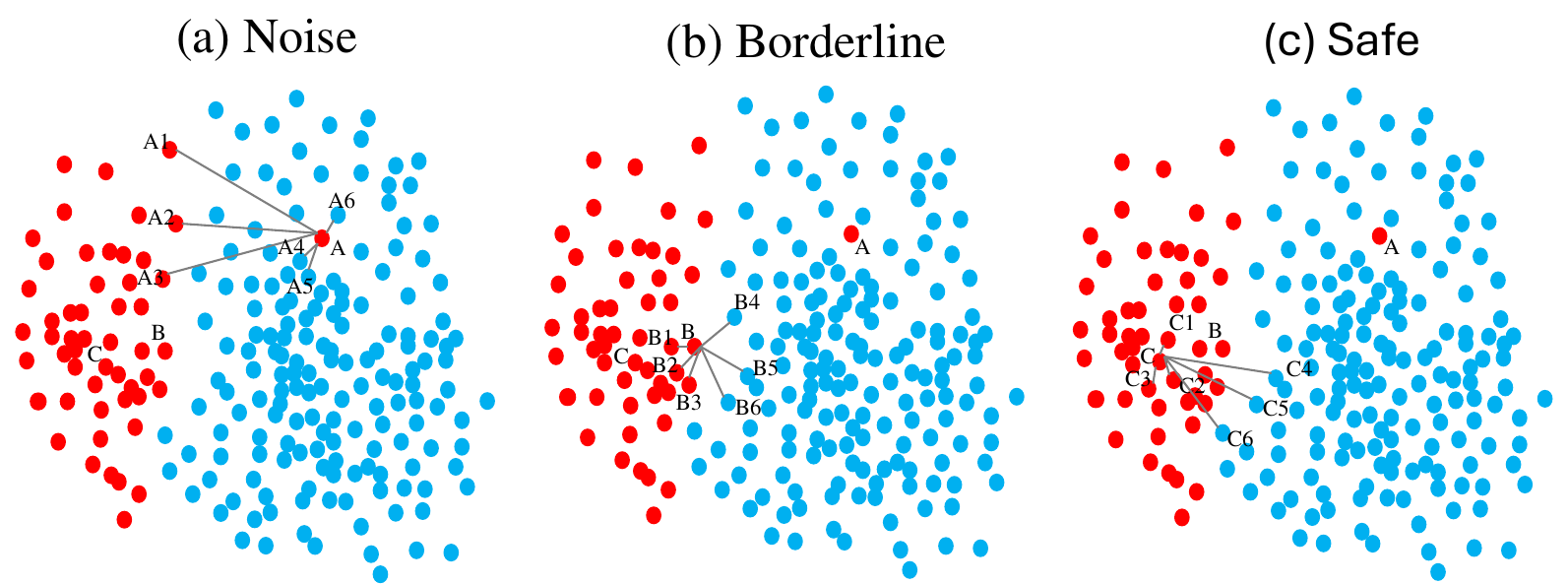} 
    \caption{: Illustration of distances between each type of minority sample and its k-nearest majority/minority neighbors.}
    \label{figure 2}
\end{figure}

A key motivation behind the GK-SMOTE algorithm is the observation that the spatial distribution of minority class samples significantly influences their reliability for synthetic instance generation. In particular, the variation in distances between each minority sample and its k-nearest neighbors, whether from the majority or minority class forms the foundation for adaptive classification into three categories: noisy, borderline, and safe. As illustrated in Figure~\ref{figure 2}, minority class samples (red points) are evaluated based on their neighborhood structure in relation to the majority class (blue points), using Gaussian Kernel Density Estimation (KDE) with \( k = 3 \). In the noise case shown in Figure~\ref{figure 2}(a), the minority sample \( A \) is surrounded predominantly by majority samples (A4, A5, A6), and is relatively distant from its own class neighbors (A1, A2, A3). KDE assigns such points a low density score, identifying them as unreliable for oversampling due to potential label noise and high overlap with the majority class. In contrast, safe samples like \( C \), shown in Figure~\ref{figure 2}(c), are enclosed by minority samples (C1, C2, C3) and distant from majority instances (C4, C5, C6). KDE assigns high probability density to such points, making them highly suitable for generating synthetic samples. Borderline samples, such as \( B \) in Figure~\ref{figure 2}(b), lie near the class decision boundary and exhibit a mixed neighborhood (majority neighbors: B4, B5, B6; minority neighbors: B1, B2, B3). These receive intermediate KDE density values, reflecting their transitional position. Unlike traditional methods that rely on distance-based relative density, GK-SMOTE uses KDE probability to systematically and adaptively categorize samples. It prioritizes synthetic generation in safe, high-density regions while minimizing oversampling in noisy, low-density regions. This KDE-driven sampling not only enhances class separability but also mitigates the risk of overfitting and label noise. As demonstrated in Table~\ref{table: 2}, this algorithmic strategy allows GK-SMOTE to dynamically adjust sampling behavior across varying data distributions, making it a self-adaptive and noise-resilient framework for imbalanced classification problems.
\vspace{-8mm}

\vspace{-7mm}

\textbf{\textit{Definition 1.}} 
Given a dataset \( D \subseteq \mathbb{R}^d \) and a data point \( p \in D \), the function \( d(p, q) \) represents the Euclidean distance between points \( p \) and \( q \). Traditional distance-based methods often fail to capture the true data distribution, particularly in imbalanced datasets. To address this, Gaussian Kernel Density Estimation (KDE) is employed to estimate the probability density function (PDF) of a given sample \( p \), providing an adaptive measure of local density. The KDE-based probability density estimate for \( p \) is given by:

\begin{equation}
    \hat{f}(p) = \frac{1}{|N(p, D)|} \sum_{q \in N(p, D)} K_h (d(p, q)),
\end{equation}

where \( K_h(\cdot) \) is the Gaussian kernel function with bandwidth \( h \), and \( N(p, D) \) is the set of nearest neighbors of \( p \) in \( D \). This adaptive and probabilistic density estimation allows for a flexible characterization of local density, distinguishing between noisy, borderline, and safe samples.

\textbf{\textit{Definition 2.}} 
For a dataset \( D \subseteq \mathbb{R}^d \) consisting of a minority class \( D^+ \) and a majority class \( D^- \), we define homogeneous and heterogeneous neighbors for any minority sample \( p \in D^+ \) using KDE-based density estimates. Homogeneous neighbors are minority samples that lie in high-density regions relative to \( p \), while heterogeneous neighbors are majority samples that reduce the density estimate of \( p \). These neighbors are defined as:

\begin{align}
HON(p) = \{ q \mid q \in D^+, \hat{f}(q) \geq \hat{f}(p) \}, \\
HEN(p) = \{ q \mid q \in D^-, \hat{f}(q) > \hat{f}(p) \},
\end{align}
\vspace{-5mm}

This classification helps identify safe and noisy regions in the dataset for effective oversampling.

\textbf{\textit{Definition 3.}} 
After estimating the KDE-based probability densities, minority samples are categorized adaptively into safe, borderline, and noisy samples based on their KDE density values. Noisy samples are those with a low \( \hat{f}(p) \), indicating their presence in sparse regions or overlapping areas with the majority class. Borderline samples are those with intermediate densities, lying near the decision boundary with a mix of homogeneous and heterogeneous neighbors. Safe samples have high \( \hat{f}(p) \), indicating clear separation from the majority class. GK-SMOTE adapts to these classifications through three steps:

\begin{itemize}
    \item {Noise Filtering with KDE:} Noisy samples with low KDE values are removed before oversampling.
    \item {Density-Based Clustering:} KDE-driven clustering replaces traditional methods to separate safe and borderline samples without requiring fixed parameters.
    \item {Density-Adaptive Synthetic Sample Generation:} Synthetic samples are generated based on the KDE probability distribution, ensuring placement in high-density regions.
\end{itemize}

\begin{table}[H]
\centering
\caption{GK-SMOTE Algorithm}
\footnotesize
        \scriptsize  
        \renewcommand{\arraystretch}{0.9}  
        \setlength{\tabcolsep}{0.6pt}  
\begin{tabular}{l }
\hline
            \textbf{Input:} Training set \( D = P \cup Q \), where \( P \) is the minority samples, \( Q \) is the majority samples. \\
            \quad \( IR \): The expected imbalance ratio. \\
            \quad \( k \): Number of nearest neighbors. \\
            \textbf{Output:} \( Syn \): Synthesized minority samples. \\
            \hline
            \( P' \gets \emptyset, \quad N \gets \left\lfloor \frac{|Q|}{IR} \right\rfloor - |P| \) \hfill // Total number of minority samples to be synthesized \\[0.5em]
                        \textbf{// Step 1: Noise Filtering and Density Calculation} \\
            \quad For each minority sample \( x_i \in P \): \\
            \quad \quad Compute \( k \) nearest neighbors; \\
            \quad \quad Count number \( m \) of majority samples in \( k \) nearest neighbors; \\
            \quad \quad If \( m = k \) then: \\
            \quad \quad \quad Filter \( x_i \) from \( P \); \\
            \quad \quad Else: \\
            \quad \quad \quad \( P' \gets P' \cup \{x_i\} \); \\
            \quad \quad \quad \( RD(x_i) \gets \) Calculate KDE-based density of \( x_i \); \\[0.5em]
            
            \textbf{// Step 2: Clustering Minority Samples} \\
            \quad Apply 2-means clustering on \( RD(P') \) to obtain clusters \( C_A \) and \( C_B \); \\
            \quad Assign \( P_A, P_B \) as minority samples corresponding to \( C_A \) and \( C_B \), respectively; \\
            \quad \( Syn \gets \emptyset \); \\[0.5em]
            
            \textbf{// Step 3: Density-Adaptive Sample Generation} \\
            \quad For each cluster \( P_j \), \( j \in \{A,B\} \): \\
            \quad \quad \( N_j \gets \frac{|P_j|}{|P|} \cdot N \); \hfill // Number of minority samples to be synthesized in \( P_j \) \\ 
            \quad For each sample \( x_i \in P_j \): \\
            \quad \quad \( w_i \gets \frac{k - w_i}{k} \); \\
            \quad \quad \( w_i \gets \frac{w_i}{\sum_{i} w_i} \); \\
            \quad \quad \( N_i \gets w_i \cdot N_j \); \\
            \quad \quad Populate(\(N_i, x, narray\)); \\[0.5em]
            
            \( Syn_j \gets Syn_j \cup \) generated samples by each minority sample in \( P_j \); \\
            return \( Syn = Syn_A \cup Syn_B \); \\
\hline
\end{tabular}
\label{table: 2}
\end{table}

\subsection{Computational Complexity Analysis} 
To analyze the computational efficiency of GK-SMOTE, we consider three key operations. First, the KDE Probability Estimation dynamically estimates density for all samples with a complexity of \( \mathcal{O}(nN) \), where \( n \) is the number of minority samples and \( N \) is the total dataset size. Second, the Density-Based Clustering applies KDE-driven clustering on the density vector with a complexity of \( \mathcal{O}(nt) \), where \( t \) is the number of iterations for convergence. Third, the Adaptive Synthetic Sample Generation generates synthetic samples based on KDE density, with a complexity of \( \mathcal{O}(ns) \), where \( s \) is the number of synthetic samples. Therefore, the overall computational complexity of GK-SMOTE is \( \mathcal{O}(nN + nt + ns) \), demonstrating its efficiency and scalability across different imbalanced datasets.

\section{Experiments}
\subsection{Dataset Selection} 
We evaluate the effectiveness of the proposed GK-SMOTE method using 27 real-world datasets from the UCI Machine Learning Repository \footnote{\url{http://archive.ics.uci.edu/ml/index.php}}, employed for binary classification tasks. Nine datasets are intrinsically imbalanced (breast\_tissue2, ecoli, heart, iris, libra, pima, segment, vehicle, and wine), while the remaining 18 datasets are artificially imbalanced by increasing the imbalance ratio (IR), defined as:
\[IR = \frac{\# \text{Majority samples}}{\# \text{Minority samples}}.\]

Each dataset is split into 75\% for training and 25\% for testing. The majority and minority samples are partitioned separately to address class imbalance. To assess GK-SMOTE's robustness in imbalanced learning with label noise, controlled levels of noise (10\%, 20\%, and 30\%) are introduced, where minority class labels are randomly flipped to the majority class, and vice versa, based on the noise rate \( c \). This process alters the class distribution in both training and testing sets.

\begin{table}[H]
\begin{flushleft}  
   \begin{minipage}{0.9\textwidth}  
        \centering
        \caption{Description of UCI Datasets}
        \scriptsize  
        \renewcommand{\arraystretch}{0.9}  
        \setlength{\tabcolsep}{0.9pt}  
   \begin{tabular}{l c c c c c|c c c c c c}
\hline
      Data & \#Dim & \#Sample & \#Maj & \#Min & IR & Data & \#Dim & \#Sample & \#Maj & \#Min & IR \\
      \hline
      breast\_tissue2 & 9 & 88 & 70 & 18 & 3.89 & symguide3 & 22 & 1041 & 947 & 94 & 10.07 \\
      ecoli & 8 & 336 & 284 & 52 & 5.46 & splice & 60 & 620 & 517 & 51 & 10.14 \\
      heart & 13 & 270 & 150 & 120 & 1.25 & mushrooms & 112 & 4628 & 4208 & 420 & 10.02 \\
      iris & 4 & 150 & 100 & 50 & 2 & isolet5 & 617 & 858 & 780 & 78 & 10 \\
      libra & 90 & 360 & 288 & 72 & 4 & codrna & 8 & 41674 & 39690 & 3969 & 10 \\
      pima & 8 & 768 & 500 & 268 & 1.87 & avila & 10 & 11695 & 10632 & 1063 & 10 \\
      segment & 16 & 2310 & 1980 & 330 & 6 & letter & 16 & 11057 & 10052 & 1005 & 10 \\
      vehicle & 18 & 846 & 647 & 199 & 3.25 & susy & 18 & 23752 & 21593 & 2159 & 10 \\
      wine & 13 & 178 & 107 & 71 & 1.51 & mocap & 33 & 48023 & 46852 & 4685 & 10 \\
      symguide1 & 4 & 4400 & 4000 & 400 & 10 & poker & 10 & 49804 & 48828 & 4882 & 10 \\
      breastcancer & 10 & 532 & 444 & 44 & 10.09 & nomao & 119 & 25852 & 24621 & 2462 & 10 \\
      creditApproval & 15 & 459 & 383 & 38 & 10.08 & magic & 10 & 13665 & 12332 & 1233 & 10 \\
      votes & 16 & 320 & 267 & 26 & 10.27 & skin & 3 & 196139 & 194198 & 1941 & 100.05 \\
      & & & & & & Online\_Retail & 7 & 541909 & 531285 & 10624 & 50.01 \\
      \hline
         \end{tabular}
         \label{table 3}
     \end{minipage}
\end{flushleft}
\end{table}

\subsection{Assessment Metrics} 
\vspace{-3mm}

To provide a more balanced evaluation, we use three metrics: Matthews Correlation Coefficient (MCC), Balanced Accuracy (BAc), and Area Under the Precision-Recall Curve (AUPRC), as shown in Eqs. \ref{eq2}-\ref{eq4}. These metrics evaluate both class-wise performance and the classifier's ability to predict minority class instances effectively.

\begin{equation}
MCC = \frac{tp \times tn - fp \times fn}{\sqrt{(tp+fp)(tp+fn)(tn+fp)(tn+fn)}},
\label{eq2}
\end{equation}
\vspace{-8mm}

\begin{equation}
Balance\ Accuracy\ (BAc) = \frac{1}{2} \left( \frac{tp}{tp+fn} + \frac{tn}{tn+fp} \right),
\label{eq3}
\end{equation}
\vspace{-10mm}

\begin{equation}
AUPRC = \int_0^1 P(R) , dR,
\label{eq4}
\end{equation}

MCC considers all confusion matrix components, Balanced Accuracy (BAc) adjusts for imbalance, AUPRC focuses on the minority class. 

\subsection{Comparison Methods and Experiment Setup} 
\vspace{-3mm}

This method introduces noise evenly across both classes, with slight IR variations due to rounding. We compare oversampling techniques' effectiveness across various models using default hyperparameters. SMOTE, Borderline-SMOTE, ADASYN, and AB-SMOTE use \(k_{\text{nn}} \in \{3, 5, 20\}\), while k-means SMOTE includes parameters like \(k \in \{2, 20, 50, 100\}\) and interpolation ratio (\(irt \in \{1, 1.5\}\)). GMF-SMOTE and NGOS include Gaussian mixture model (\(gmm \in \{0.1, 0.5, 1.0\}\)) and interpolation ratio (\(irt\)). MGD, HGDO, and GDO incorporate Gaussian distribution weights (\(gd \in \{0.1, 0.5, 1.0\}\)) and hypergraph identification weights (\(hg \in \{1.0, 2.0, 3.0\}\)). GK-SMOTE uses a KDE-based approach to adaptively generate synthetic samples based on local density variations, eliminating the need for fixed parameter tuning.
\vspace{-3mm}

\subsection{Basic Classifier Selection}
\vspace{-5mm}

We evaluated five classifiers from the scikit-learn library such as Random Forest (RF), LightGBM, Logistic Regression (LR), KNN, and Decision Tree (DT) on imbalanced datasets using 10-fold cross-validation. These classifiers were tested with SMOTE, Borderline-SMOTE, and others oversampling method inducing GK-SMOTE. Performance was compared using MCC, Balanced Accuracy (BAc), and AUPRC to identify the best classifier for handling imbalanced data with GK-SMOTE.

\begin{figure}[H]
    \centering
    \includegraphics[width=1.1\textwidth]{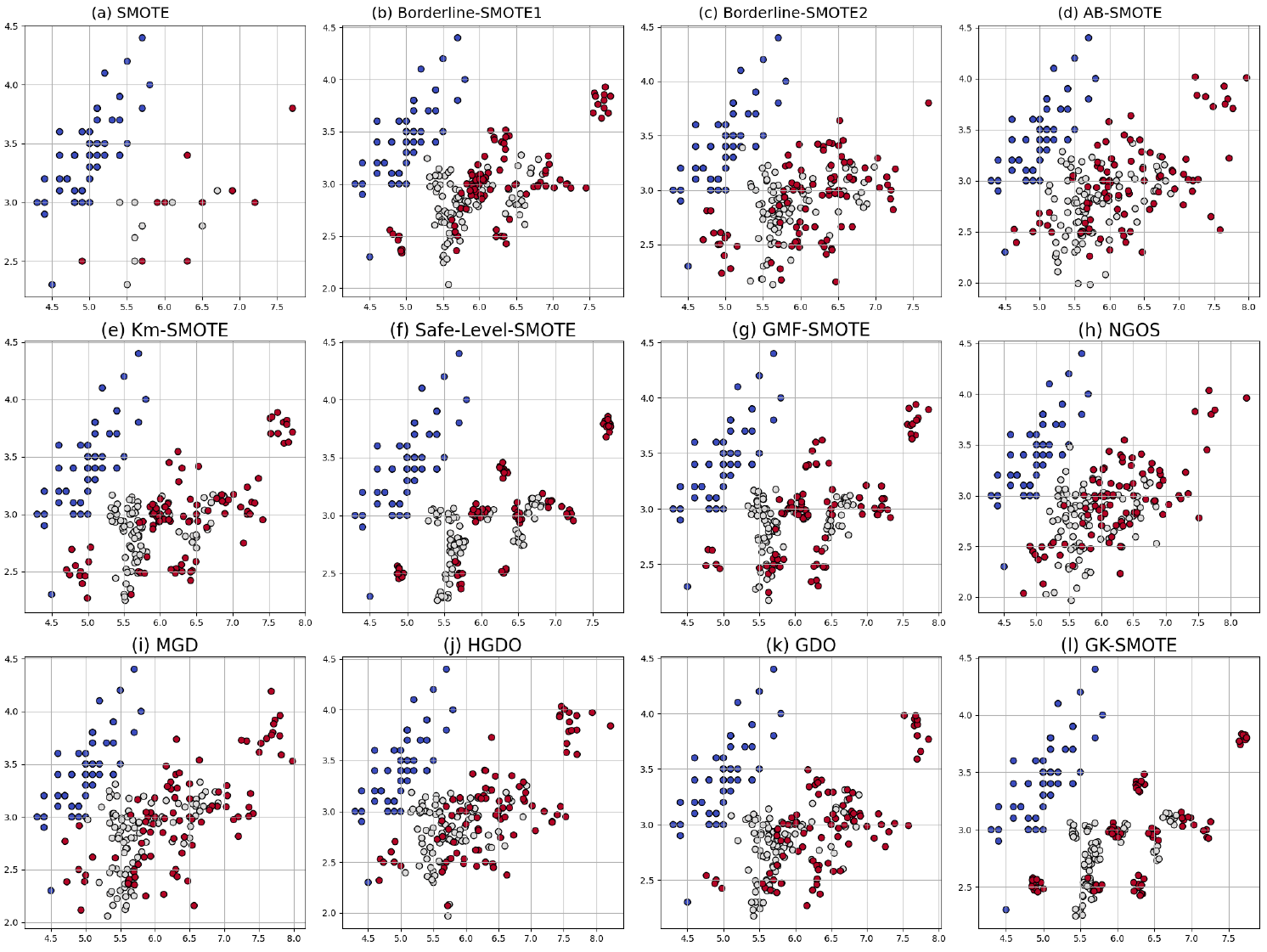} 
    \caption{Visual comparison of synthetic sample generation by various oversampling methods, including GK-SMOTE}
    \label{figure 3}
\end{figure}

\section{Results}
\subsection{Visualization Comparison on Synthetic Dataset}
To evaluate the effectiveness of various oversampling techniques, we generated scatter plots visualizing the synthetic data produced by SMOTE, Borderline-SMOTE (SMOTE1 and SMOTE2), AB-SMOTE, Km-SMOTE, Safe-Level-SMOTE, GMF-SMOTE, NGOS, MGD, HGDO, GDO, and the proposed GK-SMOTE, as illustrated in Figure~\ref{figure 3}. In each plot, blue, red, and gray dots represent majority, minority, and synthetic samples, respectively. The visualizations show that GMF-SMOTE and GK-SMOTE generate well-separated synthetic samples that maintain class balance and reduce overlap. While methods like Borderline-SMOTE and AB-SMOTE tend to generate samples near decision boundaries potentially enhancing classification in borderline regions, they also introduce a higher risk of overlap. In contrast, GK-SMOTE consistently produces more distinct, well-clustered class regions, effectively avoiding noisy samples and highlighting its superiority in preserving class separability for improved classification performance in imbalanced settings.

\subsection{Experimental Comparison and Results Analysis}
\begin{figure}[h]
    \centering
    \includegraphics[width=1.0\textwidth]{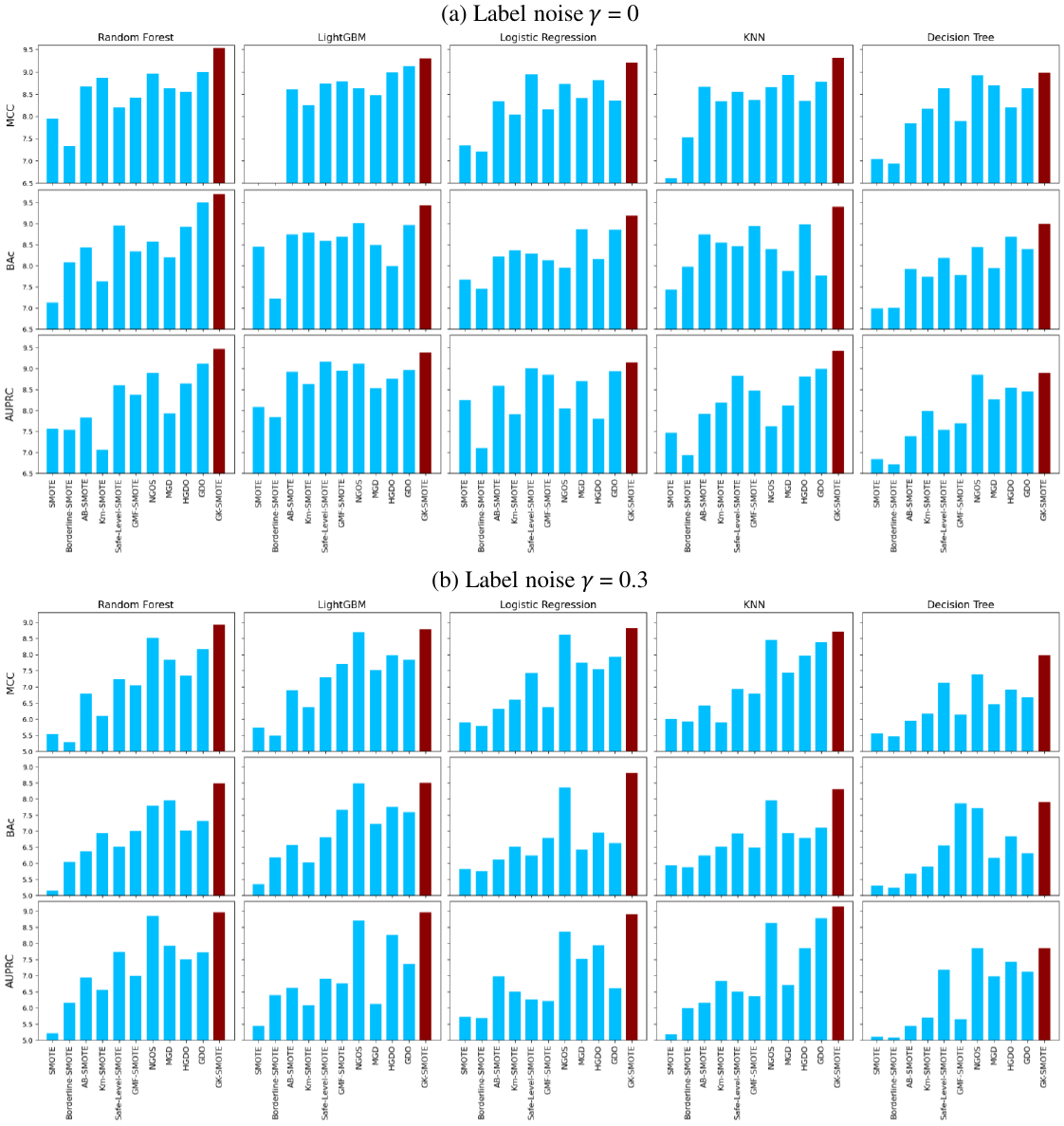} 
    \caption{Bar plots comparing of oversampling methods across classifiers under varying label noise levels such as (a) no label noise $\gamma = 0$ and (b) label noise $\gamma = 0.3$.}
    \label{figure 4}
\end{figure}

\textbf{Results under No Label Noise ($\gamma = 0$):} Under noise-free conditions, GK-SMOTE consistently achieved the best performance across all classifiers and evaluation metrics. For the MCC metric, GK-SMOTE significantly outperformed SMOTE across every classifier, with improvements of 19.9\% for Random Forest, 28.4\% for LightGBM, 25.4\% for Logistic Regression, 41.0\% for KNN, and 27.4\% for Decision Tree. These results highlight GK-SMOTE’s strong ability to enhance the classifiers’ discriminative power in distinguishing between classes in imbalanced datasets. In terms of BAc, GK-SMOTE also demonstrated substantial gains over SMOTE, improving performance by 35.9\% for Random Forest, 30.5\% for LightGBM, 37.7\% for Logistic Regression, 31.5\% for KNN, and 27.3\% for Decision Tree, indicating its effectiveness in balancing classification performance across majority and minority classes. Furthermore, for the AUPRC, GK-SMOTE achieved enhancements of 25.0\% for Random Forest, 20.8\% for LightGBM, 26.3\% for Logistic Regression, 24.5\% for KNN, and 26.3\% for Decision Tree, suggesting improved sensitivity and precision in identifying minority class instances. These findings, as shown in Figure~\ref{figure 4}(a) and Table~\ref{table 4} (Tab a), confirm GK-SMOTE’s superior performance under clean-label conditions, consistently outperforming both traditional and advanced oversampling methods across all classifiers and evaluation metrics.

\textbf{Results under Label Noise ($\gamma = 0.3$):} Even in the presence of 30\% label noise, GK-SMOTE maintained top-tier performance across all classifiers and evaluation metrics. For the MCC) metric, GK-SMOTE showed substantial improvements over SMOTE, achieving gains of 61.3\% for Random Forest, 53.6\% for LightGBM, 32.7\% for Logistic Regression, 57.0\% for KNN, and 55.2\% for Decision Tree. These results reflect GK-SMOTE’s robustness and stability in handling noisy data environments. Regarding BAc, GK-SMOTE again demonstrated superior performance with improvements of 64.9\% for Random Forest, 58.9\% for LightGBM, 47.4\% for Logistic Regression, 53.1\% for KNN, and 49.1\% for Decision Tree, indicating its strong ability to maintain balanced classification outcomes under label noise. In terms of AUPRC, GK-SMOTE outperformed all other methods, delivering enhancements of 70.9\% for Random Forest, 64.7\% for LightGBM, 66.2\% for Logistic Regression, 61.2\% for KNN, and 52.5\% for Decision Tree. These AUPRC improvements highlight GK-SMOTE’s effectiveness in preserving both precision and recall for minority classes, even under noisy conditions. As illustrated in Figure~\ref{figure 4}(b) and Table~\ref{table 4} (Tab b), GK-SMOTE continues to outperform all baseline and state-of-the-art oversampling techniques, reinforcing its robustness and adaptability in complex, real-world imbalanced classification scenarios.

\textbf{Dataset Performance Analysis:} Among the 27 datasets evaluated in our study, GK-SMOTE achieved superior performance on 21 datasets under both noise-free ($\gamma = 0$) and noisy ($\gamma = 0.3$) conditions. These datasets included many with high imbalance ratios (IR), which traditionally present greater classification challenges. GK-SMOTE’s consistent performance across these datasets highlights its generalization and robustness in handling imbalanced learning scenarios.

\begin{table}[h]
 \begin{minipage}{1.1\textwidth}
    \centering
    \caption{Comparing the performance of eleven oversampling methods, including the proposed GK-SMOTE, across five classifiers under two conditions: (Tab a) no label noise $\gamma = 0$ and (Tab b) label noise $\gamma = 0.3$. The evaluation metrics include MCC, Balanced Accuracy (BAc), and AUPRC. Here AUC: AUPRC; GBM: LightGBM; SMT: SMOTE; BSM: Borderline-SMOTE; ASM: AB-SMOTE; KMS: Km-SMOTE; SSM: Safe-Level SMOTE; GSM: GMF-SMOTE; NGO: NGOS; HGD: HGDO; GKS: GK-SMOTE}
    \label{table 4}
            \scriptsize  
        \renewcommand{\arraystretch}{1.1}  
        \setlength{\tabcolsep}{1.1pt}  
    \begin{tabular}{c c c c c c c c c c c c c c c c c c}
         \hline
         &  &  &  &  &  &  &  &  & Tab a &  &  &  &  &  & &  &  \\ 
         &  &  & MCC &  &  &  &  &  & BAc &  &  &  &  &  & AUC &  &  \\ \hline
         & RF & GBM & LR & KNN & DT &  & RF & GBM & LR & KNN & DT &  & RF & GBM & LR & KNN & DT \\ 
        SMT & 7.957 & 6.213 & 7.346 & 6.611 & 7.045 &  & 7.132 & 8.456 & 7.675 & 7.428 & 6.995 &  & 7.564 & 8.078 & 8.244 & 7.471 & 6.843 \\ 
        BSM & 7.334 & 6.072 & 7.216 & 7.525 & 6.944 &  & 8.085 & 7.228 & 7.451 & 7.978 & 6.996 &  & 7.538 & 7.835 & 7.102 & 6.934 & 6.712 \\ 
        ASM & 8.672 & 8.612 & 8.338 & 8.664 & 7.845 &  & 8.427 & 8.749 & 8.219 & 8.748 & 7.926 &  & 7.825 & 8.923 & 8.583 & 7.925 & 7.381 \\ 
        KSM & 8.868 & 8.245 & 8.041 & 8.336 & 8.171 &  & 7.625 & 8.788 & 8.365 & 8.545 & 7.748 &  & 7.063 & 8.626 & 7.911 & 8.188 & 7.994 \\ 
        SSM & 8.194 & 8.736 & 8.945 & 8.552 & 8.634 &  & 8.959 & 8.597 & 8.289 & 8.469 & 8.187 &  & 8.598 & 9.167 & 9.001 & 8.824 & 7.531 \\ 
        GSM & 8.429 & 8.786 & 8.165 & 8.367 & 7.901 &  & 8.339 & 8.693 & 8.125 & 8.942 & 7.778 &  & 8.372 & 8.942 & 8.849 & 8.476 & 7.693 \\ 
        NGO & 8.957 & 8.634 & 8.725 & 8.654 & 8.921 &  & 8.568 & 9.019 & 7.955 & 8.397 & 8.438 &  & 8.897 & 9.121 & 8.051 & 7.614 & 8.856 \\ 
        MGD & 8.636 & 8.476 & 8.417 & 8.933 & 8.698 &  & 8.195 & 8.488 & 8.865 & 7.875 & 7.939 &  & 7.931 & 8.528 & 8.696 & 8.115 & 8.269 \\ 
        HGD & 8.551 & 8.989 & 8.821 & 8.348 & 8.198 &  & 8.925 & 7.985 & 8.168 & 8.982 & 8.685 &  & 8.642 & 8.763 & 7.803 & 8.799 & 8.537 \\ 
        GDO & 8.998 & 9.118 & 8.354 & 8.778 & 8.631 &  & 9.497 & 8.967 & 8.845 & 7.765 & 8.393 &  & 9.115 & 8.961 & 8.927 & 8.991 & 8.452 \\ 
        GKS & 9.538 & 9.301 & 9.214 & 9.313 & 8.976 &  & 9.698 & 9.429 & 9.191 & 9.399 & 8.991 &  & 9.461 & 9.388 & 9.146 & 9.421 & 8.897 \\ \hline

      \hline
         &  &  &  &  &  &  &  &  & Tab b &  &  &  &  &  & &  &  \\ 
         &  &  & MCC &  &  &  &  &  & BAc &  &  &  &  &  & AUC &  &  \\ \hline
         & RF & GBM & LR & KNN & DT &  & RF & GBM & LR & KNN & DT &  & RF & GBM & LR & KNN & DT \\  
        SMT & 5.531 & 5.732 & 5.892 & 5.998 & 5.542 &  & 5.152 & 5.344 & 5.814 & 5.937 & 5.302 &  & 5.208 & 5.445 & 5.732 & 5.185 & 5.112 \\ 
        BSM & 5.292 & 5.482 & 5.784 & 5.931 & 5.453 &  & 6.043 & 6.188 & 5.763 & 5.882 & 5.239 &  & 6.157 & 6.389 & 5.681 & 5.982 & 5.077 \\ 
        ASM & 6.794 & 6.899 & 6.319 & 6.422 & 5.963 &  & 6.372 & 6.567 & 6.126 & 6.246 & 5.679 &  & 6.935 & 6.629 & 6.971 & 6.151 & 5.438 \\ 
        KSM & 6.092 & 6.372 & 6.611 & 5.891 & 6.169 &  & 6.948 & 6.032 & 6.514 & 6.518 & 5.899 &  & 6.563 & 6.073 & 6.502 & 6.842 & 5.693 \\ 
        SSM & 7.241 & 7.286 & 7.421 & 6.919 & 7.128 &  & 6.522 & 6.803 & 6.246 & 6.927 & 6.541 &  & 7.731 & 6.896 & 6.263 & 6.509 & 7.183 \\ 
        GSM & 7.055 & 7.716 & 6.362 & 6.783 & 6.145 &  & 6.996 & 7.663 & 6.788 & 6.487 & 7.871 &  & 6.983 & 6.754 & 6.21 & 6.366 & 5.65 \\ 
        NGO & 8.514 & 8.688 & 8.622 & 8.461 & 7.384 &  & 7.787 & 8.489 & 8.351 & 7.958 & 7.715 &  & 8.853 & 8.698 & 8.371 & 8.624 & 7.848 \\ 
        MGD & 7.841 & 7.514 & 7.751 & 7.436 & 6.459 &  & 7.951 & 7.225 & 6.432 & 6.946 & 6.172 &  & 7.921 & 6.124 & 7.523 & 6.711 & 6.965 \\ 
        HGD & 7.355 & 7.992 & 7.544 & 7.973 & 6.917 &  & 7.014 & 7.763 & 6.953 & 6.787 & 6.837 &  & 7.507 & 8.257 & 7.944 & 7.853 & 7.429 \\ 
        GDO & 8.165 & 7.835 & 7.926 & 8.387 & 6.686 &  & 7.316 & 7.586 & 6.624 & 7.102 & 6.314 &  & 7.712 & 7.372 & 6.614 & 8.779 & 7.118 \\ 
        GKS & 8.921 & 8.775 & 8.815 & 8.702 & 7.994 &  & 8.486 & 8.497 & 8.799 & 8.312 & 7.902 &  & 8.957 & 8.951 & 8.892 & 9.132 & 7.849 \\ \hline
    \end{tabular}
    \end{minipage}
\end{table}
\vspace{-8mm}

\subsection{Statistical Analysis}
\begin{figure}[h]
    \centering
    \includegraphics[width=1.1\textwidth]{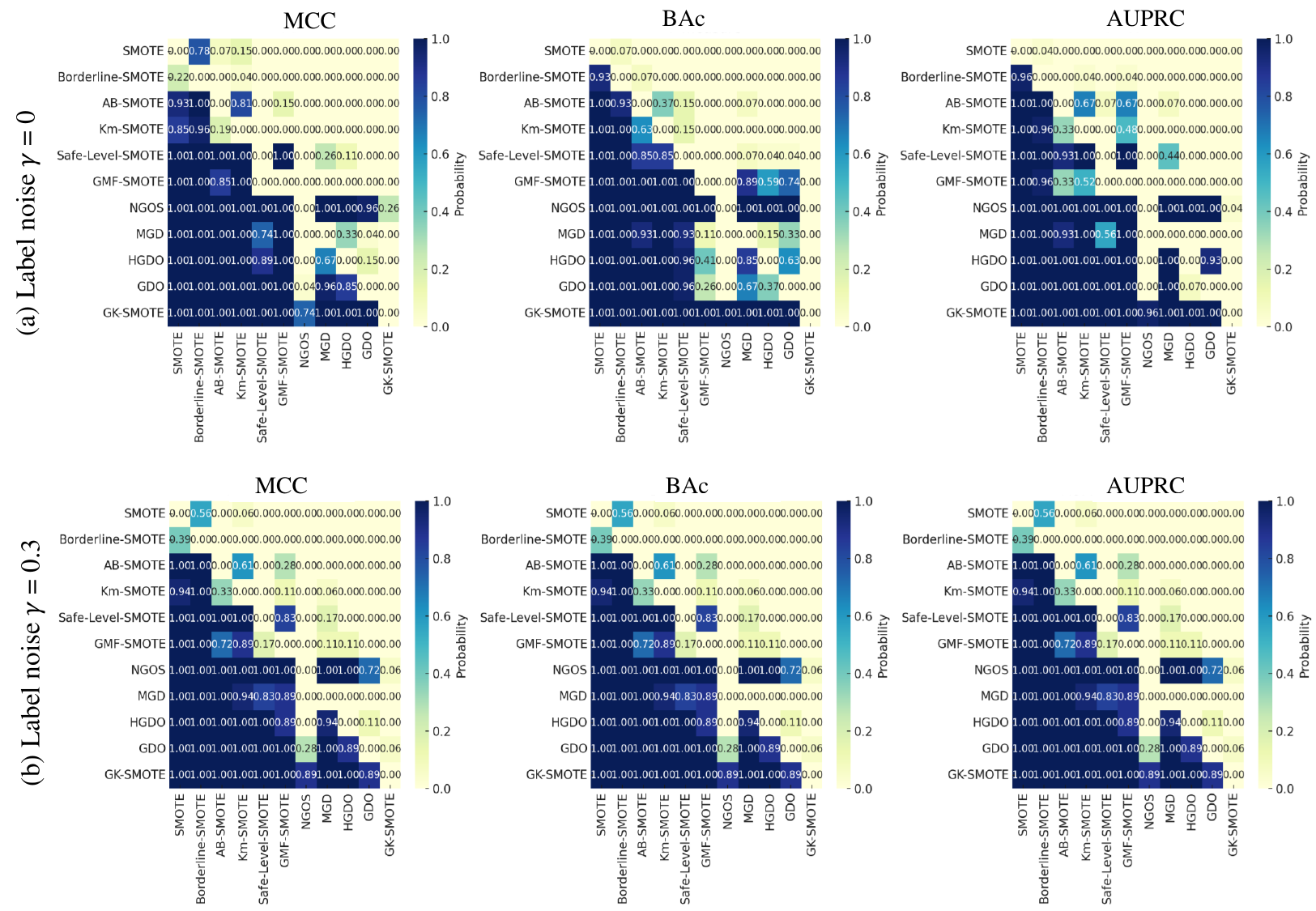} 
    \caption{Heatmaps of p-values for MCC, BAc, and AUPRC across oversampling methods under (a) no label noise $\gamma = 0$ and (b) label noise $\gamma = 0.3$ in Random Forest (RF).}
    \label{figure 5}
\end{figure}

To rigorously evaluate the statistical significance of the performance differences among oversampling methods, we employed a non-parametric Friedman test followed by the Nemenyi post-hoc test under two distinct conditions: (a) no label noise ($\gamma = 0$) and (b) moderate label noise ($\gamma = 0.3$). Figure~\ref{figure 5} shows p-value heatmaps for the Random Forest (RF) classifier across three key evaluation metrics: MCC, BAc, and AUPRC. Each heatmap summarizes pairwise comparisons between oversampling methods, where darker blue cells indicate statistically significant differences with $p < 0.05$. The results clearly show that under $\gamma = 0$, GK-SMOTE achieves statistically significant superiority ($p < 0.05$) over 100\% (11 out of 11) of the competing methods across all three metrics. When label noise increases to $\gamma = 0.3$, GK-SMOTE retains this significance for 100\% of comparisons in MCC and 90.9\% (10 out of 11) in both BAc and AUPRC, further confirming its robustness. These statistical findings reinforce that GK-SMOTE consistently performs better than other state-of-the-art and Gaussian-based oversampling techniques. These consistent results demonstrate GK-SMOTE’s resilience to label noise and its statistically validated advantage in classification performance.

\section{Conclusion}
This paper presents GK-SMOTE, a robust and hyperparameter-free oversampling method for handling imbalanced datasets with label noise. Leveraging Gaussian Kernel Density Estimation (KDE), GK-SMOTE effectively identifies dense minority regions while avoiding noisy areas, thereby enhancing class separability and mitigating noise impact. Experimental results demonstrate that GK-SMOTE consistently outperforms state-of-the-art methods, including SMOTE, Borderline-SMOTE, AB-SMOTE, km-SMOTE, and Safe-level-SMOTE, as well as modern Gaussian-based techniques such as GMF-SMOTE, NGOS, MGD, HGDO, and GDO, confirming its superior robustness, efficiency, and classification performance. Its simplicity and computational efficiency make GK-SMOTE well-suited for real-world noisy and imbalanced scenarios. In conclusion, GK-SMOTE offers an effective, user-friendly solution for improving classification accuracy without hyperparameter tuning. Future work may explore its extension to multi-class imbalances and broader model applications.
\\
\\
\noindent\textbf{Acknowledgment.} This work was supported by the National Natural Science Foundation of China under Grant 62272073.


%
%
%
%

\end{document}